\documentclass[final]{l4dc2026}

\title[WOMBET: World Model-Based Experience Transfer]{WOMBET: World Model-Based Experience Transfer for Robust and Sample-efficient Reinforcement Learning}

\usepackage{times}

\coltauthor{
  \Name{Mintae Kim} \Email{mintae.kim@berkeley.edu}\\
  \Name{Koushil Sreenath} \Email{koushils@berkeley.edu}\\
  \addr Hybrid Robotics, UC Berkeley \\ Berkeley, CA 94720, USA
}

\begin{document}

\maketitle


\begin{abstract}
Reinforcement learning (RL) in robotics is often limited by the cost and risk of data collection, motivating experience transfer from a source task to a target task.
Offline-to-online RL leverages prior data but typically assumes a given fixed dataset and does not address how to generate reliable data for transfer.
We propose \textit{World Model-Based Experience Transfer} (WOMBET), a framework that jointly generates and utilizes prior data.
WOMBET learns a world model in the source task and generates offline data via uncertainty-penalized planning, followed by filtering trajectories with high return and low epistemic uncertainty.
It then performs online fine-tuning in the target task using adaptive sampling between offline and online data, enabling a stable transition from prior-driven initialization to task-specific adaptation.
We show that the uncertainty-penalized objective provides a lower bound on the true return and derive a finite-sample error decomposition capturing distribution mismatch and approximation error.
Empirically, WOMBET improves sample efficiency and final performance over strong baselines on continuous control benchmarks, demonstrating the benefit of jointly optimizing data generation and transfer.
\end{abstract}


\vspace{-2.5mm}

\section{Introduction}
\vspace{-1mm}

Reinforcement learning assumes abundant interaction and frequent resets--conditions rarely satisfied in real-world robotics, where data collection is expensive and unsafe \cite{radosavovic2024real, li2024reinforcement, gupta2025estimation}.
As a result, RL methods achieve strong asymptotic performance but remain sample-inefficient.
Like biological systems that continually reuse past experience through internal world models, we seek to enable \textit{experience transfer}: leveraging data collected in a source environment to improve sample efficiency and robustness in a target environment \cite{wilson1994reactivation, lee2002memory}.
Existing paradigms address this only partially.
Online RL adapts to new tasks but requires extensive interaction and faces distributional shift between source and target experiences \cite{feng2023genloco, zhou2025wsrl, kim2026robust}.
Offline RL avoids unsafe exploration but but degrades under limited or mismatched data \cite{yu2020mopo, kumar2020conservative, kostrikov2021offline, yu2021combo}.
Applying it to real-world robotics often requires high-quality expert demonstrations or extensively tuned model-based controllers--assumptions that are unrealistic in many cases \cite{cai2024learning, kim2025roverfly, kim2026finite}.
Offline-to-online RL combines these by pretraining offline and fine-tuning online \cite{nair2020awac, lee2022offline, rafailov2023moto, smith2023learning, singh2020cog}, 
but assumes access to high-quality offline data without addressing how to obtain or safely reuse it. 
Model-based RL (MBRL) provides a principled way to generate data by learning a predictive world model.
Planning-based approaches use the world model for control via model predictive control (MPC) \cite{chua2018deep, wang2019exploring}, 
while Dyna-style methods use the model to generate synthetic rollouts for policy learning \cite{janner2019trust, sutton1991dyna}.
However, both faces limitations for transfer: MPC-based methods tend to generate low-diversity trajectories due to exploitation, and model-generated data can introduce bias when used without reliability control. 
Consequently, existing approaches struggle to provide reliable and adaptable data for transfer.

We propose \textit{World Model-Based Experience Transfer} (WOMBET), a unified framework that couples uncertainty-aware model-based offline data generation with online fine-tuning through iterative \textit{model-policy co-evolution}.
In each iteration, a world model generates trajectories in the source task via uncertainty-penalized MPC, and a dual criterion--high return and low epistemic uncertainty--filters reliable rollouts to form an offline dataset $D_S$.
WOMBET then performs off-policy policy optimization in the target task, to maximize a provable lower bound on the true return, while adaptive data mixing balances bias and variance between offline ($D_S$ from the source task) and online ($D_T$ collected in the target task) experience.
The world model is continuously refined using the aggregated dataset $D = D_S \cup D_T$, improving prediction accuracy and expanding the reliable planning region.
This alternating process unifies the exploitation strength of MBRL with the exploration capability of model-free RL.
Our contributions are threefold:

(1) \emph{Coupled data generation and transfer.}
We introduce a framework that jointly generates and utilizes prior data for transfer, instead of assuming a fixed offline dataset.
WOMBET uses uncertainty-aware planning to construct data and iteratively refines both the world model and policy.

(2) \emph{Reliable data curation and adaptive transfer.}
We propose (i) a dual-criterion filter selecting trajectories with high return and low epistemic uncertainty, and (ii) an adaptive sampling strategy that balances offline and online data based on TD error.

(3) \emph{Theoretical and empirical validation.}
We show that uncertainty-penalized planning maximizes a lower bound on the true return and derive a finite-sample error decomposition capturing distribution mismatch and approximation error.
Experiments demonstrate improved sample efficiency and final performance over strong baselines on continuous control benchmarks.


\begin{figure}[t!]
    \centering
    \vspace{-10mm}
    \includegraphics[width=0.95\linewidth]{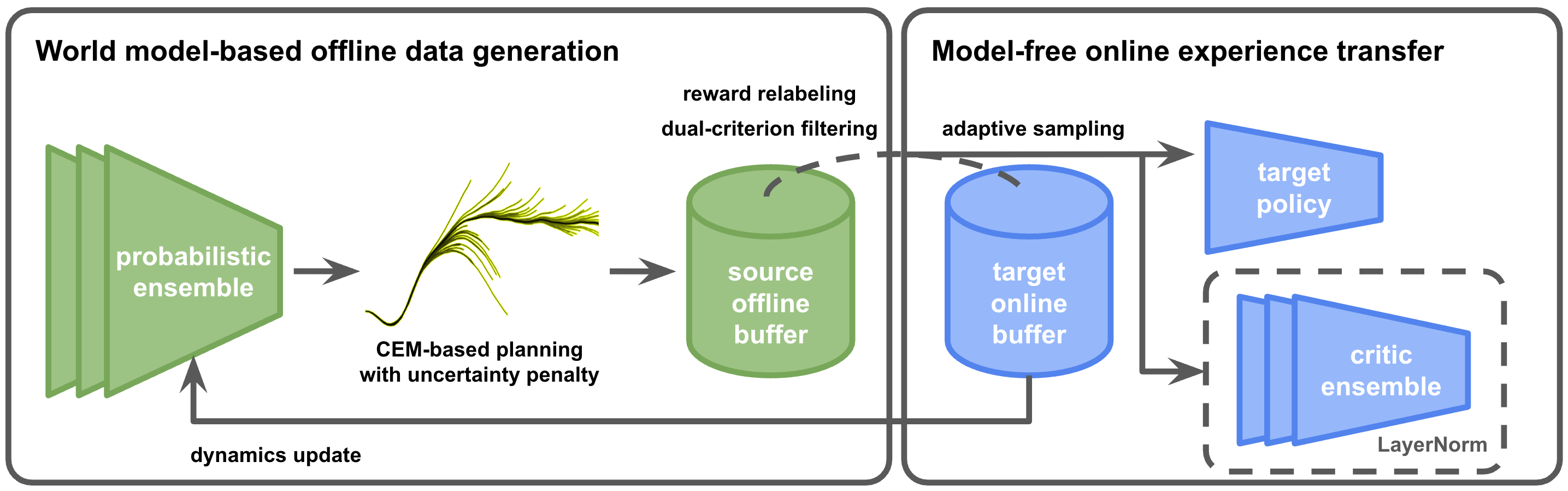}
    \vspace{-2.5mm}
    \caption{
    WOMBET pipeline: uncertainty-aware model-based data generation in the source task, followed by adaptive offline-to-online learning with iterative model updates.
    }
    \label{fig:wombet_pipeline}
    \vspace{-5mm}
\end{figure}


\vspace{-2.5mm}

\section{Related Work}
\vspace{-1mm}

\textbf{Offline-to-online RL.}
This paradigm combines offline data efficiency with online adaptability to accelerate learning when interaction is costly \cite{mao2022moore, song2022hybrid, rafailov2023moto}. 
Early methods stabilize fine-tuning via advantage-weighted updates or constrained policy improvement \cite{nair2020awac, ball2023efficient}, while later work improves robustness through fixed offline critics, ensemble critics, and transition weighting \cite{lee2022offline}. 
Architectural components such as symmetric replay buffers, ensemble critics, and layer normalization (LayerNorm) can further stabilize training without explicit regularization \cite{ba2016layer, ball2023efficient}.
Other studies explore pretraining on related tasks to enhance exploration or show that keeping the offline dataset is unnecessary when pre-trained rollouts suffice for initialization \cite{zhou2024efficient}. 
Behavior cloning and conservative value regularization are also used, but often require delicate tuning and may limit long-term performance \cite{kumar2020conservative, yu2021combo}. 
Learning-from-demonstration methods differ in relying on expert trajectories for imitation rather than addressing distributional shift during online fine-tuning \cite{hester2018deep, nair2018overcoming}. 
Overall, most methods treat offline data as static and independent of the target task, leaving data generation, critic stability, and robust exploration unresolved.

\noindent
\textbf{MBRL for data generation and offline learning.}
Model-based approaches improve sample efficiency by learning a predictive world model and leveraging it either for planning or data generation. 
Planning-based methods perform control via MPC \cite{chua2018deep, wang2019exploring}, while Dyna-style methods such as MBPO generate short synthetic rollouts to train model-free policies \cite{janner2019trust}.
MOPO introduces uncertainty-based reward penalties to mitigate model exploitation \cite{yu2020mopo}, and COMBO extends this with conservative value estimation \cite{yu2021combo}.
The main difficulty in model-based offline RL is \emph{model bias}: compounding prediction errors can degrade value estimation and lead to unsafe policies \cite{rafailov2023moto}. 
To address this, ensemble-based uncertainty penalizes unreliable or out-of-distribution (OOD) rollouts, often combined with behavior regularization or conservative value estimation. 
However, most methods use the learned model only once for rollout generation or regularization, without iterative refinement. 
WOMBET instead alternates between uncertainty-penalized data generation and online fine-tuning, updating both model and policy using aggregated experience.
This iterative process unifies model-based offline RL and online adaptation for experience transfer.

\vspace{-2.5mm}

\section{Preliminaries}
\vspace{-1mm}

We consider an MDP 
$\mathcal{M} = (\mathcal{S}, \mathcal{A}, P, r, \mu_0, \gamma)$ 
with state space $\mathcal{S}$, action space $\mathcal{A}$, transition kernel $P(s'|s,a)$, reward $r(s,a)$, initial distribution $\mu_0$, and discount factor $\gamma \in [0,1)$. 
A policy $\pi(a|s)$ aims to maximize the discounted return 
$J(\pi) = 
\mathbb{E}_{\pi, P} 
[\sum_{t=0}^\infty \gamma^t r(s_t, a_t)]$.
We study two tasks sharing dynamics but differing in reward and initial state: 
the \emph{source task} 
$\mathcal{M}_S = (\mathcal{S}, \mathcal{A}, P, r_S, \mu_0^S, \gamma)$ 
and the \emph{target task} 
$\mathcal{M}_T = (\mathcal{S}, \mathcal{A}, P, r_T, \mu_0^T, \gamma)$. 
The objective is to maximize the target return 
$J_T(\pi) = 
\mathbb{E}_{\pi, P}
[\sum_{t=0}^\infty \gamma^t r_T(s_t, a_t)]$, 
using an offline dataset from $\mathcal{M}_S$ and online data from $\mathcal{M}_T$.

\vspace{-2.5mm}

\section{WOMBET: World Model-based Experience Transfer}
\vspace{-1mm}

Unlike standard offline-to-online RL, which assumes access to offline data, WOMBET addresses the problem of \emph{generating} transferable experience.
The challenge is not only utilizing prior data, but generating reliable data from a source task that can accelerate learning in a target task.

\vspace{-2.5mm}
\subsection{World Model-based Data Generation}

A key component of WOMBET is the construction of the offline dataset $\mathcal{D}_S$ from a source task.
Rather than assuming a given dataset, WOMBET generates $\mathcal{D}_S$ via world model-based planning.

A probabilistic ensemble 
$\hat{P}_S = \{ \hat{P}_S^{(i)} \}_{i=1}^E$ 
approximates $P(s'|s,a)$ \cite{chua2018deep}. 
Under MPC, the optimal horizon-$H$ action sequence solves
\vspace{-1mm}
\begin{equation}
a_{t:t+H-1}^* =
\arg\max_{a_{t:t+H-1}}
\mathbb{E}_{\hat{P}_S}
\!\left[\sum_{k=0}^{H-1}\!\gamma^k r_S(s_{t+k}, a_{t+k})\right].
\end{equation}
Only the first action $a_t^*$ is executed. Repeating this procedure yields model-based rollouts $\mathcal{D}_S$. 

In WOMBET, this process is iterative--the model $\hat{P}$ is continually refined with $\mathcal{D} = \mathcal{D}_S \cup \mathcal{D}_T$, progressively improving predictive accuracy and uncertainty estimation over iterations.

\vspace{-2.5mm}

\subsection{Robust Planning with a Provable Performance Lower Bound}

Source experience $\mathcal D_S$ provides a strong initialization, but rollouts under the world model $\hat P$ remain prone to bias and over-optimism. To ensure reliability, WOMBET introduces an \emph{uncertainty penalty} during MPC, forming a conservative surrogate that lower-bounds the true return under dynamics $P$. The true return $J_P(\pi)=\mathbb E_{\pi,P}[\sum_{t=0}^{H-1}r(s_t,a_t)]$ deviates from the model return $J_{\hat P}(\pi)=\mathbb E_{\pi,\hat P}[\sum_{t=0}^{H-1}r(s_t,a_t)]$ due to model bias.
Let $V_P^\pi(s)=\mathbb E_{\pi,P}[\sum_{t=0}^{H-1}r(s_t,a_t)\,|\,s_0=s]$ denote the value function of MPC under true dynamics. We assume that $V_P^\pi$ is $L_v$-Lipschitz in state, i.e., $|V_P^\pi(s_1)-V_P^\pi(s_2)|\le L_v\|s_1-s_2\|$ for all $(s_1,s_2)$. We also assume that the ensemble uncertainty $u(s,a)$ upper-bounds the Wasserstein distance between transition kernels, $W_1(P,\hat P)\le u(s,a)$ for all $(s,a)$.
Define the one-step model bias $G^\pi(s,a)=\mathbb E_{\hat P}[V_P^\pi(s')]-\mathbb E_P[V_P^\pi(s')]$. Then
\vspace{-1mm}
\begin{equation}
|G^\pi(s,a)|\le L_v\,W_1(P,\hat P)\le L_v\,u(s,a).
\label{eq:onestep}
\end{equation}
This follows from the Lipschitz property of $V_P^\pi$ and the definition of $W_1$ (see \cite{yu2020mopo}).

Expanding the Bellman telescoping yields
\vspace{-1mm}
\begin{equation}
J_P(\pi)\ge\mathbb E_{\pi,\hat P}\Big[\sum_{t=0}^{H-1}(r(s_t,a_t)-\lambda\,u(s_t,a_t))\Big],
\label{eq:lb-pen}
\end{equation}
for any $\lambda\ge L_v$. Since $J_P(\pi)-J_{\hat P}(\pi)=\sum_t\mathbb E_{\pi,\hat P}[G^\pi(s_t,a_t)]$, (\ref{eq:onestep}) and bounding each term by $L_v u(s_t,a_t)$ yields (\ref{eq:lb-pen}). Defining the penalized reward $\tilde r(s,a)=r(s,a)-\lambda u(s,a)$ gives
\vspace{-1mm}
\begin{equation}
J_P(\pi)\ge\mathbb E_{\pi,\hat P}\Big[\sum_{t=0}^{H-1}\tilde r(s_t,a_t)\Big]=:\tilde J_{\hat P}(\pi),
\label{eq:simple-lb}
\end{equation}
so maximizing $\tilde J_{\hat P}$ under $\hat P$ maximizes a certified lower bound on $J_P$.
Since (\ref{eq:simple-lb}) holds at each time horizon, the receding-horizon controller that greedily maximizes the penalized return satisfies
\vspace{-1mm}
\begin{equation}
J_P(\pi_{\mathrm{MPC}})\ge\mathbb E_{\pi_{\mathrm{MPC}},\hat P}\Big[\sum_{t=0}^{H-1}\tilde r(s_t,a_t^\star)\Big],
\label{eq:mpc-lb}
\end{equation}
which is precisely the quantity optimized online. Thus, uncertainty-penalized MPC maximizes a provable lower bound on the true return, balancing reward exploitation under $\hat P$ with avoidance of high-uncertainty regions where $u(s,a)$ is large.

\vspace{-2.5mm}

\subsection{Dual-Criterion Filtering for Reliable Offline Data from MBRL}

A central component of WOMBET is the generation of the offline dataset $\mathcal{D}_S$.
Unlike prior work that assumes a given dataset, WOMBET generates $\mathcal{D}_S$ from rollouts of a source-trained world model $\hat{P}$, via model-based planning and explicitly controls its reliability through filtering.
Since $\hat{P}\!\neq\! P$, naïvely using all synthetic trajectories induces model bias. WOMBET controls dataset \emph{reliability} at generation time via MPC planning and uncertainty-aware filtering.
A probabilistic ensemble $\hat{P}=\{\hat{P}^{(i)}\}_{i=1}^E$ is used within MPC to optimize short-horizon returns under $\hat{P}$, synthesizing diverse high-return behaviors without costly real interaction \cite{chua2018deep}. Epistemic uncertainty--estimated by ensemble predictive variance--serves as a proxy for model error. Rather than correcting this error only during learning (e.g., MOPO \cite{yu2020mopo}), WOMBET \emph{preemptively} removes unreliable trajectories by a dual-criterion filter: a trajectory $\tau$ is accepted and appended to $\mathcal{D}_S$ iff
\vspace{-1mm}
\begin{equation}
\bar{u}(\tau):=\tfrac{1}{H}\sum_{t=0}^{H-1}u(s_t,a_t)\le u_{\mathrm{th}},\qquad
J(\tau):=\sum_{t=0}^{H-1}\gamma^t r_S(s_t,a_t)\ge J_{\mathrm{th}},
\end{equation}
where $u_{\mathrm{th}}$ and $J_{\mathrm{th}}$ are user-specified thresholds controlling uncertainty and return levels, respectively.
Accepted rollouts are relabeled with the target reward $r_T(s,a)$ to form $\mathcal{D}_S$. This step (i) suppresses bias by excluding high-uncertainty regions and (ii) yields a compact, high-value dataset that accelerates transfer. As online data accumulates, the model is refined and filtering is repeated, gradually expanding $\mathcal{D}_S$ toward a broader yet reliable state-action region.

\vspace{-2.5mm}

\subsection{Mixed Data Training}

Given the constructed dataset $\mathcal{D}_S$ and online data $\mathcal{D}_T$, 
WOMBET performs policy optimization using a mixture of offline and online experience.

Let $\mathcal{D}_S$ and $\mathcal{D}_T$ denote offline and online datasets collected in the source and the target tasks respectively. 
Batches are sampled from
$\mathcal{D}_{\text{mix}} = \alpha \mathcal{D}_S + (1-\alpha)\mathcal{D}_T$, 
where $\alpha \in [0,1]$ adaptively balances bias and variance. 
An ensemble critic $\{ Q_{\phi_i} \}_{i=1}^N$ is trained with a robust Bellman target:
\vspace{-1mm}
\begin{equation}
\mathcal{L}_{\text{critic}}^{(i)} =
\mathbb{E}_{(s,a,r,s') \sim \mathcal{D}_{\text{mix}}}
\big[
(Q_{\phi_i}(s,a) - \mathcal{T}^\pi Q_{\bar{\phi}}(s,a))^2
\big],
\end{equation}
where 
$\mathcal{T}^\pi Q(s,a) =
r_T(s,a) - \mathbb{I}_{s \in \mathcal{D}_S} \lambda u(s,a)
+ \gamma \mathbb{E}_{a'\!\sim\!\pi(\cdot|s')}
[\min_i Q_{\bar{\phi}_i}(s',a')]$.
Here $u(s,a)$ estimates epistemic uncertainty via a model ensemble~\cite{chua2018deep}, and $\lambda>0$ scales the uncertainty penalty, inducing a robust lower bound on the return. 
LayerNorm~\cite{ba2016layer} stabilizes the critic, and the actor maximizes the conservative value:
$\mathcal{L}_{\text{actor}} =
-\mathbb{E}_{s \sim \mathcal{D}_{\text{mix}}}
[\min_i Q_{\phi_i}(s, \pi_\phi(s))]$.

\vspace{-2.5mm}

\subsection{Bounding Extrapolation Error via Implicit Regularization}

When the policy $\pi$ starts interacting with the target environment, it encounters $(s,a)$ pairs that lie outside the support of the offline dataset $\mathcal{D}_S$ \cite{lee2020addressing, park2024value}. 
On such OOD inputs, function approximators often overestimate Q-values--\emph{extrapolation error}--causing value explosion and unstable updates.
WOMBET mitigates this not by explicit penalties but through \emph{implicit regularization} inherited from RL with prior data (RLPD)~\cite{ball2023efficient}, a SAC-based algorithm that incorporates LayerNorm, ensuring conservative and stable value estimates through two complementary mechanisms: architectural bounding and algorithmic robustness.
Each critic employs LayerNorm across hidden layers to constrain activations and thereby bound Q-value magnitudes. 
For a critic $Q_\phi(s,a) = w^\top \mathrm{ReLU}(\psi_\theta(s,a))$, where $\psi_\theta(s,a)$ is LayerNorm-activated,
\vspace{-1mm}
\begin{equation}
\|Q_\phi(s,a)\| = \|w^\top \mathrm{ReLU}(\psi_\theta(s,a))\| \le \|w\|\,\|\psi_\theta(s,a)\|.
\end{equation}
Since LayerNorm normalizes $\|\psi_\theta(s,a)\|$ across inputs, $\|Q_\phi(s,a)\|$ remains bounded by $\|w\|$, even for OOD samples. 
This architectural constraint prevents value explosion and stabilizes gradient propagation during learning.
Complementing this bound, WOMBET uses an ensemble critic $\{Q_{\phi_i}\}_{i=1}^N$ and computes Bellman targets using the ensemble minimum:
\begin{equation}
y(r,s') = r + \gamma \min_i Q_{\bar{\phi}_i}(s',a'), \qquad a' \sim \pi(\cdot|s').
\end{equation}
Taking the ensemble minimum reduces overestimation, particularly where critic disagreement indicates uncertainty. 
This ensemble-induced pessimism acts as an implicit regularizer, keeping extrapolated values conservative without extra penalty terms. 
Together, these regularizations maintain bounded and stable value estimates as WOMBET transitions from offline to online learning.


\begin{figure}[t!]
    \centering
    \vspace{-10mm}
    \includegraphics[width=0.95\linewidth]{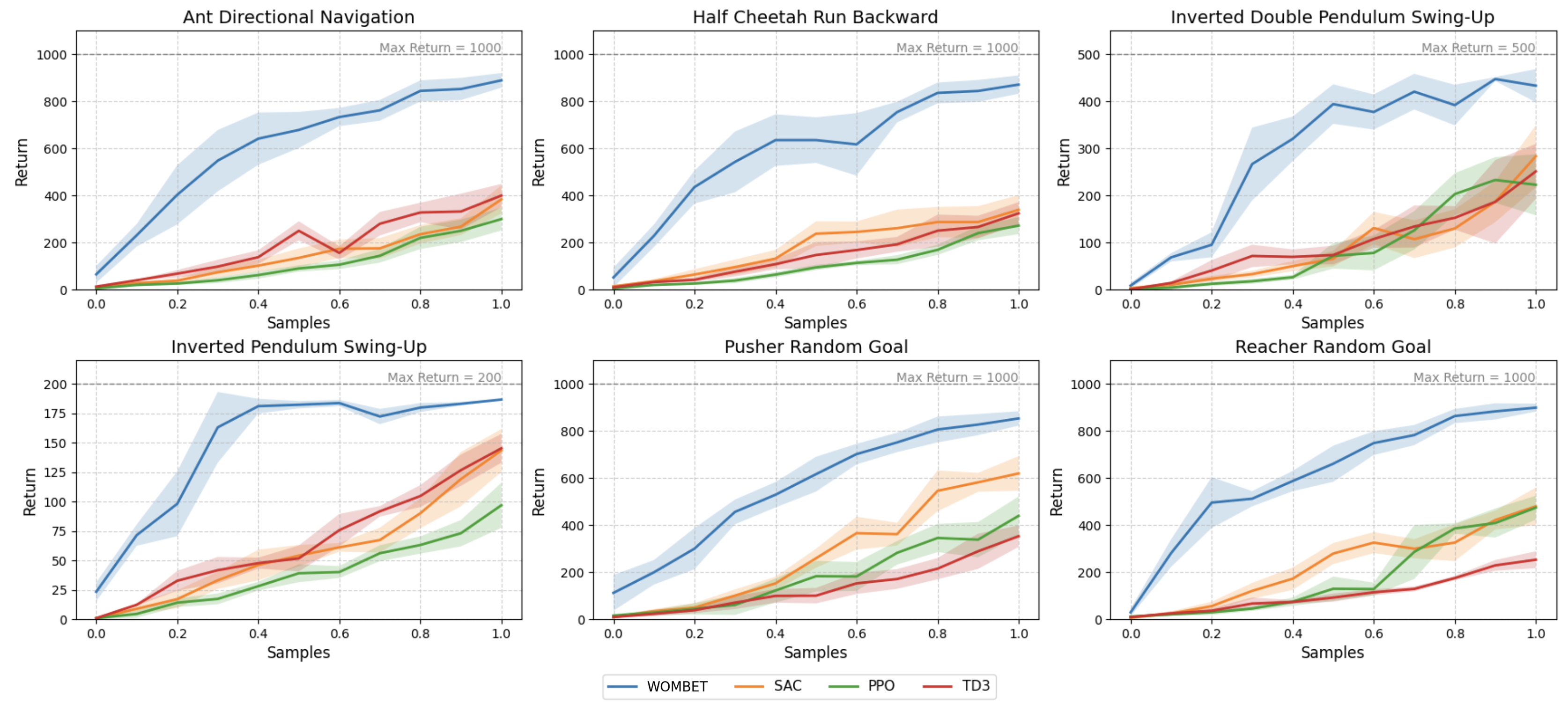}
    \vspace{-2.5mm}
    \caption{
    Sample efficiency of WOMBET vs.\ online RL baselines across target tasks.
    }
    \label{fig:wombet_vs_online_rl_all_envs}
    \vspace{-2.5mm}
\end{figure}

\vspace{-2.5mm}

\subsection{Mitigating Distributional Shift with Adaptive Sampling}

While implicit regularization limits local extrapolation error, a more fundamental issue arises from the \emph{global} shift in the state-action distribution as the policy evolves. 
As $\pi_k$ adapts, its visitation distribution $d_{\pi_k}$ diverges from the offline distribution $d_{\mathcal{D}_S}$, which biases value estimation. 
WOMBET mitigates this through \emph{adaptive sampling}, dynamically balancing bias and variance in critic updates.
At iteration $k$, critic updates draw samples from a mixture distribution $d_{\mathrm{mix}}^{(k)} = \alpha_k d_{\mathcal{D}_S} + (1 - \alpha_k) d_{\mathcal{D}_T^{(k)}}$, where $\mathcal{D}_T^{(k)}$ is the online replay buffer and $\alpha_k\!\in\![0,1]$ controls the contribution of offline and online data. 
A larger $\alpha_k$ stabilizes training using reliable offline data, while a smaller $\alpha_k$ reduces bias by emphasizing online experience. Let $\hat{Q}$ denote the learned critic that approximates the true value function $Q^{\pi_k}$. The trade-off follows the domain adaptation bound  $\mathbb{E}_{d_{\pi_k}}\!\big[|\hat{Q}-Q^{\pi_k}|\big] \le \mathbb{E}_{d_{\mathrm{mix}}^{(k)}}\!\big[|\hat{Q}-Q^{\pi_k}|\big] + L\,W_1(d_{\pi_k},d_{\mathrm{mix}}^{(k)})$, where $L$ is the Lipschitz constant of the pointwise error map $f(s,a)=|\hat{Q}(s,a)-Q^{\pi_k}(s,a)|$ with respect to $(s,a)$.
The optimal mixture coefficient $\alpha_k^*$ minimizes this upper bound, defined as $\alpha_k^* = \arg\min_{\alpha\in[0,1]} \{\epsilon_{\mathrm{approx}}\!\big(\mathcal{D}_{\mathrm{mix}}(\alpha)\big) + L\,W_1\!\big(d_{\pi_k},\mathcal{D}_{\mathrm{mix}}(\alpha)\big)\}$, where $\epsilon_{\mathrm{approx}}(\mathcal{D}) := \mathbb{E}_{d_{\mathcal{D}}}\!\big[\,|\hat Q - Q^{\pi_k}|\,\big]$ is the critic’s approximation error under dataset/distribution $\mathcal{D}$, and $\mathcal{D}_{\mathrm{mix}}(\alpha)$ denotes the mixture dataset induced by $d_{\mathrm{mix}}^{(k)} = \alpha\, d_{\mathcal{D}_S} + (1-\alpha)\, d_{\mathcal{D}_T^{(k)}}$.

WOMBET measures critic reliability using the mean absolute TD error on recent online data. 
When the TD error is large (implies underfitting), more offline samples are used to stabilize updates. 
As the TD error decreases, sampling shifts toward online data to improve adaptation. 
At iteration $k$, the TD error is $\delta_T^{(k)} = \mathbb{E}_{(s,a,r,s')\!\sim\!\mathcal{B}_T}\!\big[|Q_\phi(s,a)-y(r,s')|\big]$, where $\mathcal{B}_T$ denotes the target task replay buffer. It is smoothed via exponential averaging $\bar{\delta}_T^{(k)} = (1-\beta_{\mathrm{ema}})\bar{\delta}_T^{(k-1)} + \beta_{\mathrm{ema}}\delta_T^{(k)}$, and used to update the mixing ratio $\alpha_k = \mathrm{clip}\!\big(\lambda_{\mathrm{gain}}\bar{\delta}_T^{(k)}, \alpha_{\min}, \alpha_{\max}\big)$.
This rule approximates the gradient of the error bound with respect to $\alpha_k$, adjusting data composition based on critic uncertainty. 
As learning progresses, $\bar{\delta}_T^{(k)}$ stabilizes and $\alpha_k$ converges to a balanced ratio, keeping WOMBET stable in early training and increasingly adaptive later, enabling robust and efficient transfer learning.


\begin{figure}[t!]
    \centering
    \vspace{-10mm}
    \includegraphics[width=0.95\linewidth]{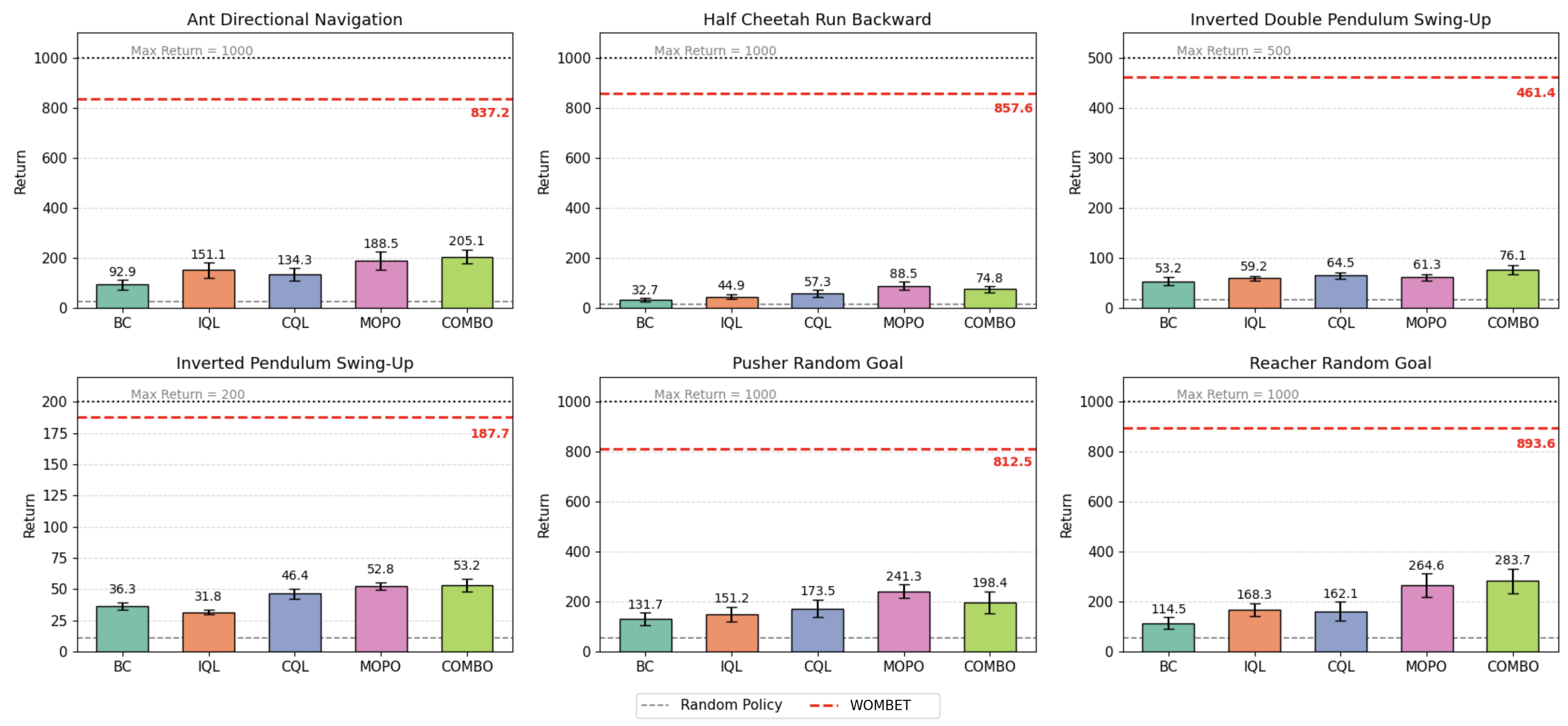}
    \vspace{-2.5mm}
    \caption{
    Comparison of offline RL baselines and WOMBET across target tasks. 
    Bars show offline performance on $\mathcal{D}_S$ and the red dashed line denotes WOMBET's post-adaptation return.
    }
    \label{fig:wombet_vs_offline_rl}
    \vspace{-5mm}
\end{figure}

\vspace{-2.5mm}

\subsection{Robust Surrogate Optimization and Performance Guarantee}

Building on the preceding analysis, we establish a unified performance bound for WOMBET. 
Its two-layer design--\emph{uncertainty-penalized offline data generation} and \emph{adaptive fine-tuning}--ensures reliable offline initialization and stable online transfer. 
In the offline phase, MBRL with an uncertainty penalty produces conservative trajectories for $\mathcal{D}_S$, which are later relabeled with $r_T$ to align with the target task. 
The online phase is entirely model-free and optimizes unpenalized target returns from real interactions. 
The remaining suboptimality arises from how accurately the learned critic $\hat{Q}$ approximates the true target value $Q_T^\pi$ during fine-tuning.

The total estimation error can be decomposed into two sources:
\begin{equation}
\sup_{(s,a)} |Q_T^\pi(s,a) - \hat{Q}(s,a)| 
\le 
\underbrace{|Q_T^\pi - Q_{\mathrm{mix}}^\pi|}_{\text{(a) Distribution mismatch}}
+
\underbrace{|Q_{\mathrm{mix}}^\pi - \hat{Q}|}_{\text{(b) Finite-sample approximation}},
\end{equation}
where $Q_{\mathrm{mix}}^\pi$ denotes the ideal action-value under the mixed visitation distribution $d_{\mathrm{mix}}=\alpha\,d_{\mathcal{D}_S}+(1-\alpha)\,d_{\mathcal{D}_T}$. 
WOMBET minimizes both components through its model-based data generation and adaptive sampling mechanisms.

Given the mixed dataset $\mathcal{D}_{\mathrm{mix}}$ and critic class $\mathcal{F}$, 
a PAC-style~\cite{agarwal2019reinforcement} bound gives
\begin{equation}
|Q_{\mathrm{mix}}^\pi(s,a)-\hat{Q}(s,a)|
\le
\mathcal{O}\!\left(
\tfrac{r_{\max}}{1-\gamma}
\sqrt{\tfrac{\ln(|\mathcal{F}|/\delta)}{|\mathcal{D}_{\mathrm{mix}}|}}
\right),
\end{equation}
where $r_{\max}:=\sup_{s,a}|r_T(s,a)|$, $\gamma\in(0,1)$ is the discount factor, and $\delta\in(0,1)$ is the confidence parameter.
Intuitively, PAC bounds state that with probability at least $1-\delta$, the critic’s generalization error decreases as the dataset size grows. 
WOMBET mitigates this term via (i) \emph{adaptive sampling}, which balances bias and variance, and (ii) \emph{implicit regularization} (LayerNorm and ensemble-min targets), which improve generalization and prevent value divergence.

Deviation between the mixed-data critic and the true target critic satisfies
\begin{equation}
|Q_T^\pi(s,a)-Q_{\mathrm{mix}}^\pi(s,a)|
\le
\tfrac{\gamma}{1-\gamma}L_V\Delta_P,
\end{equation}
where $\Delta_P:=\sup_{s,a}W_1\!\big(P_T(\cdot|s,a),P_S(\cdot|s,a)\big)$ 
quantifies the dynamics discrepancy between target and source, 
and $L_V$ is the Lipschitz constant of the value function $V_P^\pi$ with respect to state. 
WOMBET reduces this term through \emph{dual-criterion filtering}, which constrains $\mathcal{D}_S$ to high-return, low-uncertainty trajectories. 
As online data accumulates and the mixing weight $\alpha_k$ decreases, the influence of mismatched source samples further diminishes.

Combining both bounds, WOMBET optimizes a robust surrogate objective while explicitly controlling approximation and distributional errors. 
The resulting policy satisfies
\begin{equation}
J_T(\pi_{\mathrm{WOMBET}}) 
\ge
\tilde{J}_T(\pi_{\mathrm{WOMBET}})
-
\mathcal{O}\!\left(
\tfrac{\gamma}{1-\gamma}L_V\Delta_P
+
\tfrac{r_{\max}}{1-\gamma}
\sqrt{\tfrac{\ln(|\mathcal{F}|/\delta)}{|\mathcal{D}_{\mathrm{mix}}|}}
\right),
\end{equation}
showing that WOMBET achieves conservative yet asymptotically consistent transfer by coupling model-based uncertainty-aware data generation with model-free fine-tuning.


\begin{figure}[t!]
    \centering
    \vspace{-10mm}
    \includegraphics[width=0.95\textwidth]{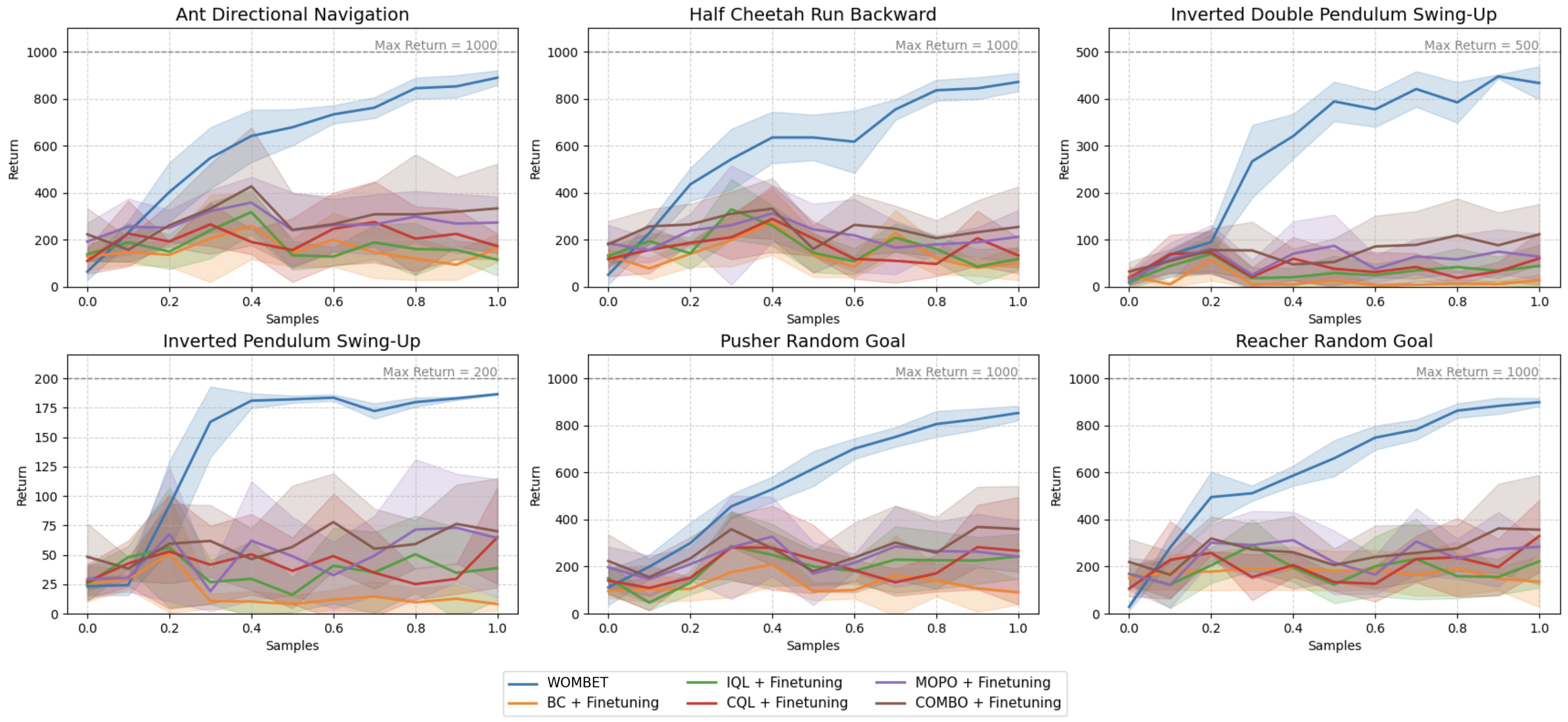}
    \vspace{-2.5mm}
    \caption{
    WOMBET vs.\ offline-to-online baselines (offline RL with fine-tuning) across target tasks.
    }
    \label{fig:wombet_vs_offline_ft}
    \vspace{-5mm}
\end{figure}


\vspace{-2.5mm}

\section{Experiments}

We evaluate WOMBET on diverse MuJoCo benchmarks to examine:  
(1) \textbf{sample efficiency}: whether experience transfer accelerates learning over online RL trained from scratch,
(2) \textbf{necessity of fine-tuning}: whether transferred data alone can solve the target task without fine-tuning,
(3) \textbf{transfer effectiveness}: how WOMBET compares with existing offline-to-online baselines, and
(4) \textbf{component contribution}: the impact of dual-criterion filtering and adaptive sampling.
We measure normalized return and shaded regions indicate standard deviation across 5 seeds.

\vspace{-2.5mm}

\subsection{Sample Efficiency: WOMBET vs.\ Online RL}

We evaluate whether WOMBET improves sample efficiency over standard online RL.
We compare against SAC, PPO, and TD3 trained from scratch on the target task $\mathcal{M}_T$ \cite{schulman2017ppo, haarnoja2018sac, fujimoto2018td3}.
As shown in Figure~\ref{fig:wombet_vs_online_rl_all_envs}, WOMBET learns faster and achieves higher asymptotic returns.
The filtered dataset $\mathcal{D}_S$ provides a strong prior, allowing WOMBET to focus on refinement rather than initial exploration.
This gain comes from uncertainty-aware planning and filtering, which produce reliable and transferable experience.
In contrast, SAC, PPO, and TD3 rely on uninformed exploration, resulting in slower convergence.
Across six environments, WOMBET attains comparable or better final returns with less than half the interaction budget, demonstrating improved sample efficiency.

\vspace{-2.5mm}

\subsection{Necessity of Online Fine-tuning: WOMBET vs.\ Offline RL}

We evaluate the necessity of online fine-tuning by comparing WOMBET to BC, IQL, CQL, MOPO, and COMBO, each trained only on the model-generated dataset $\mathcal{D}_S$ \cite{kostrikov2021offline, kumar2020conservative, yu2020mopo, yu2021combo}. 
These methods are evaluated on the target task $\mathcal{M}_T$ without further interaction, forming a zero-shot transfer setting.
As shown in Figure~\ref{fig:wombet_vs_offline_rl}, offline-only methods perform well when $\mathcal{M}_S$ and $\mathcal{M}_T$ are closely aligned but degrade under moderate shifts. 
Although $\mathcal{D}_S$ is reliable, it reflects source dynamics and rewards rather than those of $\mathcal{M}_T$. 
Without online updates, policies cannot adapt to new rewards or unseen state-action regions.
WOMBET addresses this by refining both policy and model through interaction, enabling adaptation beyond the offline data support. 
The consistent gap shows that gains come not only from pretraining but from effectively leveraging prior data for exploration and refinement.


\begin{figure}[t!]
    \centering
    \vspace{-10mm}
    \includegraphics[width=\textwidth]{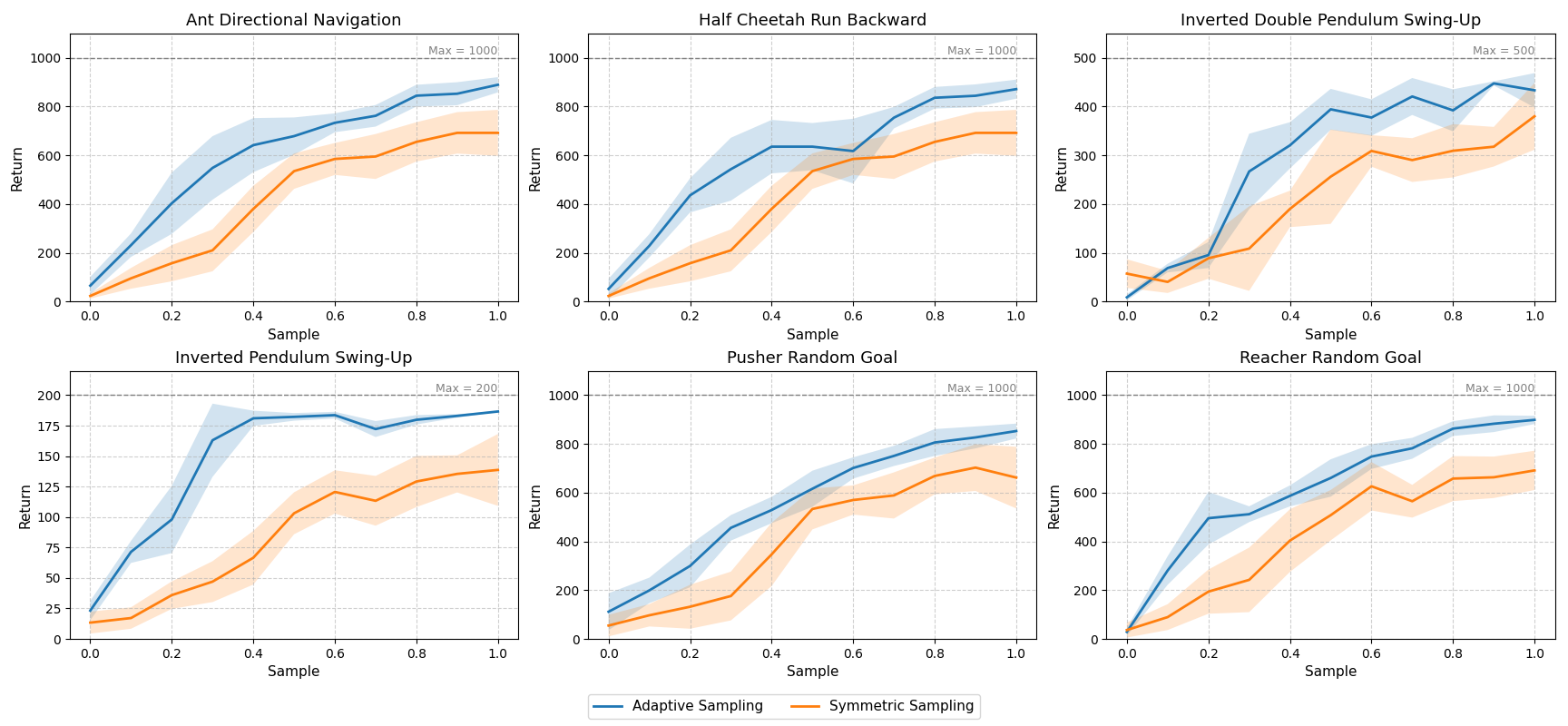}
    \vspace{-5mm}
    \caption{
    Comparison between WOMBET with adaptive sampling (blue) and a symmetric fixed-ratio variant (orange) across target tasks.
    }
    \label{fig:adaptive_vs_symmetric}
    \vspace{-5mm}
\end{figure}

\vspace{-2.5mm}

\subsection{Effectiveness of Experience Transfer: WOMBET vs.\ Offline-to-Online Baselines}

We next compare WOMBET with state-of-the-art offline-to-online RL methods to evaluate its transfer effectiveness. 
Each baseline is pretrained offline on the same model-generated dataset $\mathcal{D}_S$ and then fine-tuned online in the target environment $\mathcal{M}_T$.
Note that all methods are initialized with the same dataset $\mathcal{D}_S$ generated by WOMBET, which isolates the effect of the transfer mechanism while controlling for data construction.
As shown in Figure~\ref{fig:wombet_vs_offline_ft}, WOMBET consistently achieves higher sample efficiency and asymptotic return than all baselines. 
This improvement stems from its integrated transfer design: dual-criterion filtering provides reliable initialization, and adaptive sampling enables a smooth offline-to-online transition, maintaining a balanced bias-variance trade-off from the outset. 
In contrast, conventional pretrain-finetune pipelines face abrupt distributional shifts--their offline-trained critics and actors often fail to adapt to online data, causing unstable or inefficient exploration. 
By coupling model-based pretraining with robust online transfer, WOMBET avoids this mismatch and achieves stable, data-efficient policy improvement, confirming its ability to transfer model-based experience into high-performing online control.

\vspace{-2.5mm}

\subsection{Importance of WOMBET’s Core Components for Offline Data Generation}

\begin{figure}[t!]
    \centering
    \vspace{-10mm}
    \includegraphics[width=\textwidth]{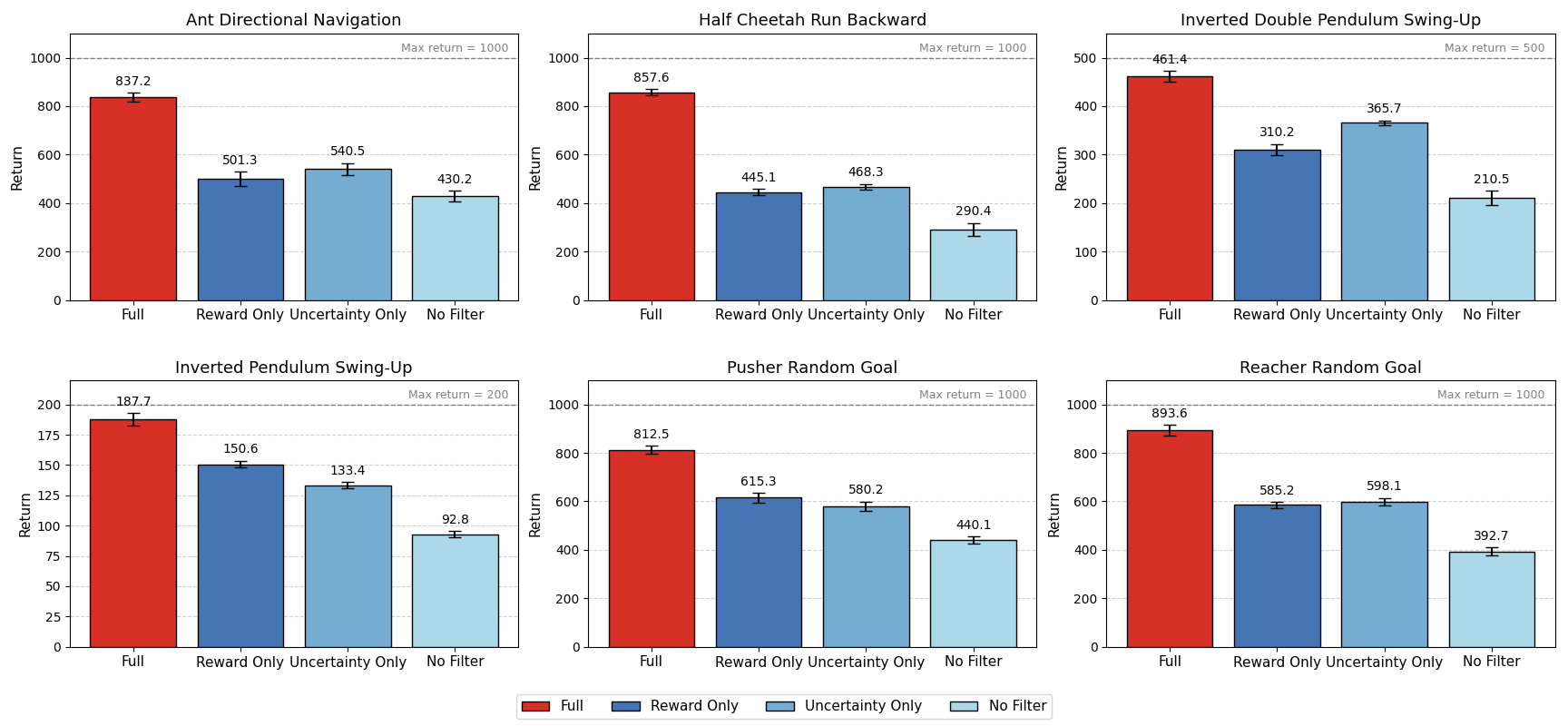}
    \vspace{-5mm}
    \caption{
    Ablation of WOMBET's dual-criterion filter across target tasks.
    }
    \label{fig:dual_criterion}
\end{figure}

We evaluate whether WOMBET’s performance arises from its key components: adaptive sampling and dual-criterion filtering.
For adaptive sampling, we compare the full method--where $\alpha_k$ is adjusted using the critic’s TD error--to a fixed-ratio baseline ($\alpha_k=0.5$).
As shown in Figure~\ref{fig:adaptive_vs_symmetric}, the fixed schedule is stable but learns slower and reaches lower returns. 
The adaptive rule shifts weight from offline to on-policy data, maintaining a bias-variance balance during training.
For dual-criterion filtering, we compare the full rule (high reward and low uncertainty) with reward-only, uncertainty-only, and unfiltered variants (Figure~\ref{fig:dual_criterion}).
All degraded variants reduce performance: reward-only includes unreliable samples, uncertainty-only is overly conservative, and no filtering amplifies model bias. 
The dual-criterion filter yields trajectories that are both reliable and task-relevant. 
Across tasks, these components are complementary and necessary for stable transfer and data-efficient fine-tuning.


\vspace{-2.5mm}

\section{Discussion and Conclusion}

WOMBET presents a unified framework for offline-to-online RL that combines conservative model-based data generation with model-free fine-tuning. 
Unlike prior methods that assume a fixed offline dataset, WOMBET constructs transferable experience from a source task. 
It addresses a key limitation of MBRL--policy exploitation of model errors--by integrating uncertainty-aware data generation with adaptive policy updates.
In the source task, uncertainty-penalized MPC with dual-criterion filtering produces reliable, high-return trajectories that form the offline dataset. 
In the target task, model-free learning with adaptive sampling balances stability from offline data and exploration through interaction. 
Regularized updates and normalization further stabilize value estimation.
Overall, WOMBET couples conservative data generation with adaptive learning, mitigating model bias, improving sample efficiency, and enabling robust transfer across continuous control tasks.


\acks{This work was supported in part by NSF CMMI-2140650 and in part by Design of Robustly Implementable Autonomous and Intelligent Machines, DARPA award number HR00112490425.}
\bibliography{reference}

@article{schulman2017ppo,
  title        = {Proximal Policy Optimization Algorithms},
  author       = {Schulman, John and Wolski, Filip and Dhariwal, Prafulla and Radford, Alec and Klimov, Oleg},
  journal      = {arXiv preprint arXiv:1707.06347},
  year         = {2017}
}

@inproceedings{haarnoja2018sac,
  title        = {Soft Actor-Critic: Off-Policy Maximum Entropy Deep Reinforcement Learning with a Stochastic Actor},
  author       = {Haarnoja, Tuomas and Zhou, Aurick and Abbeel, Pieter and Levine, Sergey},
  booktitle    = {Proceedings of the 35th International Conference on Machine Learning},
  series       = {Proceedings of Machine Learning Research},
  volume       = {80},
  pages        = {1861--1870},
  year         = {2018},
  publisher    = {PMLR}
}

@inproceedings{fujimoto2018td3,
  title        = {Addressing Function Approximation Error in Actor-Critic Methods},
  author       = {Fujimoto, Scott and van Hoof, Herke and Meger, David},
  booktitle    = {International Conference on Machine Learning},
  pages        = {1582--1591},
  year         = {2018}
}

@article{agarwal2019reinforcement,
  title={Reinforcement learning: Theory and algorithms},
  author={Agarwal, Alekh and Jiang, Nan and Kakade, Sham M and Sun, Wen},
  journal={CS Dept., UW Seattle, Seattle, WA, USA, Tech. Rep},
  volume={32},
  pages={96},
  year={2019}
}

@misc{kim2026robust,
      title={Robust Adversarial Policy Optimization Under Dynamics Uncertainty}, 
      author={Mintae Kim and Koushil Sreenath},
      year={2026},
      eprint={2604.10974},
      archivePrefix={arXiv},
      primaryClass={cs.LG},
      url={https://arxiv.org/abs/2604.10974}, 
}

@article{kim2026finite,
  title={Finite Memory Belief Approximation for Optimal Control in Partially Observable Markov Decision Processes},
  author={Kim, Mintae},
  journal={arXiv preprint arXiv:2601.03132},
  year={2026}
}

@inproceedings{zhou2025wsrl,
  title        = {Efficient Online Reinforcement Learning Fine-Tuning Need Not Retain Offline Data},
  author       = {Zhou, Zhiyuan and Peng, Andy and Li, Qiyang and Levine, Sergey and Kumar, Aviral},
  booktitle    = {ICLR},
  year         = {2025},
  note         = {arXiv preprint arXiv:2412.07762},
  archivePrefix= {arXiv},
  eprint       = {2412.07762}
}

@inproceedings{ball2023efficient,
  title={Efficient online reinforcement learning with offline data},
  author={Ball, Philip J and Smith, Laura and Kostrikov, Ilya and Levine, Sergey},
  booktitle={International Conference on Machine Learning},
  pages={1577--1594},
  year={2023},
  organization={PMLR}
}

@article{kostrikov2021offline,
  title={Offline reinforcement learning with implicit q-learning},
  author={Kostrikov, Ilya and Nair, Ashvin and Levine, Sergey},
  journal={arXiv preprint arXiv:2110.06169},
  year={2021}
}

@article{kumar2020conservative,
  title={Conservative q-learning for offline reinforcement learning},
  author={Kumar, Aviral and Zhou, Aurick and Tucker, George and Levine, Sergey},
  journal={Advances in neural information processing systems},
  volume={33},
  pages={1179--1191},
  year={2020}
}

@article{yu2021combo,
  title={Combo: Conservative offline model-based policy optimization},
  author={Yu, Tianhe and Kumar, Aviral and Rafailov, Rafael and Rajeswaran, Aravind and Levine, Sergey and Finn, Chelsea},
  journal={Advances in neural information processing systems},
  volume={34},
  pages={28954--28967},
  year={2021}
}

@article{yu2020mopo,
  title={Mopo: Model-based offline policy optimization},
  author={Yu, Tianhe and Thomas, Garrett and Yu, Lantao and Ermon, Stefano and Zou, James Y and Levine, Sergey and Finn, Chelsea and Ma, Tengyu},
  journal={Advances in Neural Information Processing Systems},
  volume={33},
  pages={14129--14142},
  year={2020}
}

@article{zhou2024efficient,
  title={Efficient online reinforcement learning fine-tuning need not retain offline data},
  author={Zhou, Zhiyuan and Peng, Andy and Li, Qiyang and Levine, Sergey and Kumar, Aviral},
  journal={arXiv preprint arXiv:2412.07762},
  year={2024}
}

@article{park2024value,
  title={Is value learning really the main bottleneck in offline RL?},
  author={Park, Seohong and Frans, Kevin and Levine, Sergey and Kumar, Aviral},
  journal={Advances in Neural Information Processing Systems},
  volume={37},
  pages={79029--79056},
  year={2024}
}

@article{nair2020awac,
  title={Awac: Accelerating online reinforcement learning with offline datasets},
  author={Nair, Ashvin and Gupta, Abhishek and Dalal, Murtaza and Levine, Sergey},
  journal={arXiv preprint arXiv:2006.09359},
  year={2020}
}

@inproceedings{lee2022offline,
  title={Offline-to-online reinforcement learning via balanced replay and pessimistic q-ensemble},
  author={Lee, Seunghyun and Seo, Younggyo and Lee, Kimin and Abbeel, Pieter and Shin, Jinwoo},
  booktitle={Conference on Robot Learning},
  pages={1702--1712},
  year={2022},
  organization={PMLR}
}

@inproceedings{rafailov2023moto,
  title={Moto: Offline to online fine-tuning for model-based reinforcement learning},
  author={Rafailov, Rafael and Hatch, Kyle Beltran and Kolev, Victor and Martin, John D and Phielipp, Mariano and Finn, Chelsea},
  booktitle={Workshop on Reincarnating Reinforcement Learning at ICLR 2023},
  year={2023}
}

@inproceedings{hester2018deep,
  title={Deep q-learning from demonstrations},
  author={Hester, Todd and Vecerik, Matej and Pietquin, Olivier and Lanctot, Marc and Schaul, Tom and Piot, Bilal and Horgan, Dan and Quan, John and Sendonaris, Andrew and Osband, Ian and others},
  booktitle={Proceedings of the AAAI conference on artificial intelligence},
  volume={32},
  number={1},
  year={2018}
}

@inproceedings{nair2018overcoming,
  title={Overcoming exploration in reinforcement learning with demonstrations},
  author={Nair, Ashvin and McGrew, Bob and Andrychowicz, Marcin and Zaremba, Wojciech and Abbeel, Pieter},
  booktitle={2018 IEEE international conference on robotics and automation (ICRA)},
  pages={6292--6299},
  year={2018},
  organization={IEEE}
}

@article{lee2020addressing,
  title={Addressing distribution shift in online reinforcement learning with offline datasets},
  author={Lee, Seunghyun and Seo, Younggyo and Lee, Kimin and Abbeel, Pieter and Shin, Jinwoo},
  year={2020}
}

@article{mao2022moore,
  title={MOORe: Model-based offline-to-online reinforcement learning},
  author={Mao, Yihuan and Wang, Chao and Wang, Bin and Zhang, Chongjie},
  journal={arXiv preprint arXiv:2201.10070},
  year={2022}
}

@article{song2022hybrid,
  title={Hybrid rl: Using both offline and online data can make rl efficient},
  author={Song, Yuda and Zhou, Yifei and Sekhari, Ayush and Bagnell, J Andrew and Krishnamurthy, Akshay and Sun, Wen},
  journal={arXiv preprint arXiv:2210.06718},
  year={2022}
}

@article{singh2020cog,
  title={Cog: Connecting new skills to past experience with offline reinforcement learning},
  author={Singh, Avi and Yu, Albert and Yang, Jonathan and Zhang, Jesse and Kumar, Aviral and Levine, Sergey},
  journal={arXiv preprint arXiv:2010.14500},
  year={2020}
}

@article{chua2018deep,
  title={Deep reinforcement learning in a handful of trials using probabilistic dynamics models},
  author={Chua, Kurtland and Calandra, Roberto and McAllister, Rowan and Levine, Sergey},
  journal={Advances in neural information processing systems},
  volume={31},
  year={2018}
}

@article{janner2019trust,
  title={When to trust your model: Model-based policy optimization},
  author={Janner, Michael and Fu, Justin and Zhang, Marvin and Levine, Sergey},
  journal={Advances in neural information processing systems},
  volume={32},
  year={2019}
}

@article{wang2019exploring,
  title={Exploring model-based planning with policy networks},
  author={Wang, Tingwu and Ba, Jimmy},
  journal={arXiv preprint arXiv:1906.08649},
  year={2019}
}

@article{sutton1991dyna,
  title={Dyna, an integrated architecture for learning, planning, and reacting},
  author={Sutton, Richard S},
  journal={ACM Sigart Bulletin},
  volume={2},
  number={4},
  pages={160--163},
  year={1991},
  publisher={ACM New York, NY, USA}
}

@article{li2024reinforcement,
    author = {Zhongyu Li and Xue Bin Peng and Pieter Abbeel and Sergey Levine and Glen Berseth and Koushil Sreenath},
    title ={Reinforcement learning for versatile, dynamic, and robust bipedal locomotion control},
    journal = {The International Journal of Robotics Research},
    volume = {0},
    number = {0},
    pages = {02783649241285161},
    year = {2025},
    doi = {10.1177/02783649241285161},
    URL = {https://doi.org/10.1177/02783649241285161},
    eprint = {https://doi.org/10.1177/02783649241285161}
}

@misc{cai2024learning,
      title={Learning-based Trajectory Tracking for Bird-inspired Flapping-Wing Robots}, 
      author={Cai, Jiaze and Sangli, Vishnu and Kim, Mintae and  Sreenath, Koushil},
      year={2024},
      eprint={2411.15130},
      archivePrefix={arXiv},
      primaryClass={cs.RO},
      url={https://arxiv.org/abs/2411.15130}, 
}

@article{radosavovic2024real,
  title={Real-world humanoid locomotion with reinforcement learning},
  author={Radosavovic, Ilija and Xiao, Tete and Zhang, Bike and Darrell, Trevor and Malik, Jitendra and Sreenath, Koushil},
  journal={Science Robotics},
  volume={9},
  number={89},
  pages={eadi9579},
  year={2024},
  publisher={American Association for the Advancement of Science}
}

@article{kim2025roverfly,
  title={RoVerFly: Robust and Versatile Implicit Hybrid Control of Quadrotor-Payload Systems},
  author={Kim, Mintae and Cai, Jiaze and Sreenath, Koushil},
  journal={arXiv preprint arXiv:2509.11149v2},
  year={2025}
}

@inproceedings{feng2023genloco,
  title={Genloco: Generalized locomotion controllers for quadrupedal robots},
  author={Feng, Gilbert and Zhang, Hongbo and Li, Zhongyu and Peng, Xue Bin and Basireddy, Bhuvan and Yue, Linzhu and Song, Zhitao and Yang, Lizhi and Liu, Yunhui and Sreenath, Koushil and others},
  booktitle={Conference on Robot Learning},
  pages={1893--1903},
  year={2023},
  organization={PMLR}
}

@article{smith2023learning,
  title={Learning and adapting agile locomotion skills by transferring experience},
  author={Smith, Laura and Kew, J Chase and Li, Tianyu and Luu, Linda and Peng, Xue Bin and Ha, Sehoon and Tan, Jie and Levine, Sergey},
  journal={arXiv preprint arXiv:2304.09834},
  year={2023}
}

@article{ba2016layer,
  title={Layer normalization},
  author={Ba, Jimmy Lei and Kiros, Jamie Ryan and Hinton, Geoffrey E},
  journal={arXiv preprint arXiv:1607.06450},
  year={2016}
}

@article{wilson1994reactivation,
  title={Reactivation of hippocampal ensemble memories during sleep},
  author={Wilson, Matthew A and McNaughton, Bruce L},
  journal={Science},
  volume={265},
  number={5172},
  pages={676--679},
  year={1994},
  publisher={American Association for the Advancement of Science}
}

@article{lee2002memory,
  title={Memory of sequential experience in the hippocampus during slow wave sleep},
  author={Lee, Albert K and Wilson, Matthew A},
  journal={Neuron},
  volume={36},
  number={6},
  pages={1183--1194},
  year={2002},
  publisher={Elsevier}
}

@inproceedings{gupta2025estimation,
  title={Estimation of aerodynamics forces in dynamic morphing wing flight},
  author={Gupta, Bibek and Kim, Mintae and Park, Albert and Sihite, Eric and Sreenath, Koushil and Ramezani, Alireza},
  booktitle={2025 IEEE/RSJ International Conference on Intelligent Robots and Systems (IROS)},
  pages={7210--7215},
  year={2025},
  organization={IEEE}
}

\end{document}